\documentclass[acmsmall,screen, final]{acmart}

\usepackage{latexsym}
\usepackage{amssymb}
\usepackage{amsmath}
\usepackage{amsthm}
\usepackage{booktabs}
\usepackage{enumitem}
\usepackage{graphicx}
\usepackage{multirow}
\usepackage{hyperref}
\usepackage{float}
\usepackage{xcolor}
\usepackage{listings}
\usepackage{lstautogobble}
\usepackage{longtable}
\usepackage{listings}
\usepackage{supertabular}

\AtBeginDocument{%
  }

\setcopyright{acmcopyright}
\copyrightyear{2024}
\acmYear{2024}
\acmDOI{XXXXXXX.XXXXXXX}

\begin{document}

\title{Towards Optimizing a Retrieval Augmented Generation using Large Language Model on Academic Data}


\author{Anum Afzal}
\email{anum.afzal@tum.de}
\author{Juraj Vladika}
\email{juraj.vladika@tum.de}
\author{Gentrit Fazlija}
\email{gentrit.fazlija@tum.de}
\author{Andrei Staradubets}
\email{ge89ped@mytum.de}
\author{Florian Matthes}
\email{matthes@tum.de}

\affiliation{%
  \institution{Technical University of Munich}
  \streetaddress{Boltzmannstr. 3}
  \city{Munich}
  \country{Germany}
}


\begin{abstract}
  Given the growing trend of many organizations integrating Retrieval Augmented Generation (RAG) into their operations, we assess RAG on domain-specific data and test state-of-the-art models across various optimization techniques. We incorporate four optimizations; Multi-Query, Child-Parent-Retriever, Ensemble Retriever, and In-Context-Learning, to enhance the functionality and performance in the academic domain. We focus on data retrieval, specifically targeting various study programs at a large technical university. We additionally introduce a novel evaluation approach, the RAG Confusion Matrix designed to assess the effectiveness of various configurations within the RAG framework. By exploring the integration of both open-source (e.g., Llama2, Mistral) and closed-source (GPT-3.5 and GPT-4) Large Language Models, we offer valuable insights into the application and optimization of RAG frameworks in domain-specific contexts. Our experiments show a significant performance increase when including multi-query in the retrieval phase.
\end{abstract}





\maketitle

\section{Introduction}
Given the recent surge of Large Language Models (LLMs), they have found application in many industrial use cases. Practitioners in various domains are increasingly eager to harness the power of LLMs in their work and operations. However, the use of LLMs in a domain-specific context is not straightforward, as these models are predominantly trained on publicly available data, making them less suited for specialized domains. This challenge led to the emergence of Retrieval Augmented Generation (RAG) frameworks, which offer a solution by enabling the integration of domain-specific data with the expansive knowledge base of LLMs, thereby enhancing their utility and relevance in various applications.

We opt for a study program use-case targeting university students. The data distribution of this use case is potentially different from what most LLMs are normally trained on, hence making it a good test bed for testing RAG performance as well as its various optimizations. The rapid development in all areas of research has given rise to many interdisciplinary and multifaceted university study programs, students are often left overwhelmed with many different programs to choose from. Understanding the prerequisites, study regulations, and curricula of programs can pose a challenge, which encourages the students to use digital solutions such as chatbots \cite{chen2023artificial}. We take advantage of this use case and create a domain-specific dataset for evaluating and optimizing RAG. For this purpose, we scrape the study program descriptions from a university website and construct a Question Answering dataset of 200 questions and answers, along with relevant context to evaluate the performance of the RAG system with its various components. Since LLMs are not trained on the study program description and requirements, this use case is ideal for RAG. Given a student query, it finds the relevant study program from the Vector Database and uses that information to answer the student's query. In this paper, we present a comprehensive evaluation of RAG and its various optimizations on a question-answering task to evaluate its performance over domain-specific data. Our main contributions are as follows:


\begin{enumerate}

    \item We constructed a dataset CurriculumQA in English and German, which was used to get the first insights into research and development of RAG pipelines for domain-specific data. 
    \item We experiment with and evaluate various RAG enhancement approaches to find the optimal configuration for an RAG system over domain-specific data.
    \item We evaluate the LLM responses based on \textit{Relevance}, \textit{Coherence}, \textit{Faithfulness}, and \textit{Fluency}. We compare the automatic and human evaluation metrics.
\end{enumerate}

\section{Related Work}

Early approaches to RAG involved relatively simple methods of retrieving relevant documents based on the query and then processing these documents to come up with a response to a given question \cite{chen-etal-2017-reading}. However, recent advancements have seen more sophisticated integration of retrieval and generation processes. Notable implementations of RAG have demonstrated the ability to dynamically retrieve and incorporate relevant information during the generation process, thereby significantly enhancing the quality and relevance of the generated text \cite{lewis2020retrieval}. These advancements have been facilitated by improvements in both the retrieval mechanisms, which have become more efficient and effective at finding relevant information, and the generative models, which have become better at integrating and contextualizing the retrieved information \cite{cai2022recent}. A recent survey \cite{gao2024retrievalaugmented} separates RAG approaches into \textit{naive RAG} and \textit{advanced RAG}. The naive RAG approach follows a traditional process that includes indexing, retrieval, and generation, which is also characterized as a “Retrieve-and-Read” framework \cite{zhu2021retrieving}. On one hand, advanced RAG introduces specific improvements to enhance the retrieval quality, by employing pre-retrieval and post-retrieval strategies. Pre-retrieval strategies include query rewriting with an LLM \cite{ma-etal-2023-query} or query expansion methods like HyDE \cite{gao-etal-2023-precise}, which generates a hypothetical response to the query first and then uses the responses to search the database. On the other hand, post-retrieval methods focus on selecting the most essential information from retrieved documents, since feeding all of them at once to the LLM can lead to information overload and noisy context. This includes reranking the retrieved documents with neural models \cite{glass-etal-2022-re2g} or summarizing the context of retrieved documents before passing them as context \cite{an2021retrievalsum}. Our study builds on top of these advancements by conducting a comprehensive evaluation of the usefulness of recent retrieval augmentations like multi-query, child-parent, and ensemble retrieval, as well as the generation technique in-context learning. 

Retrieval-augmented generation systems can be deployed for a wide variety of NLP tasks involving text generation. The most popular is open-domain question answering \cite{siriwardhana2023improving}, where the task is to answer a given question with no provided context, i.e., the system has to first search through large knowledge bases in order to find an answer. Beyond QA, RAG can also be used for generative tasks like machine translation by retrieving example sentences from a corpus in target language \cite{cai-etal-2021-neural}, or for dialogue generation by guiding the conversation with retrieved exemplar responses from previous dialogues \cite{gupta-etal-2021-controlling}. Applying AI and NLP methods to help education and students has been widely studied \cite{zhang2021ai, kasneci2023chatgpt}. Example use cases of NLP technologies for educational applications include language learning \cite{settles2018second}, grammatical error correction \cite{grundkiewicz2019neural}, and automated essay scoring \cite{ramesh2022automated}. There have also been QA systems, often based on retrieval and generation, developed and evaluated for various domains, including QA for compliance \cite{abualhaija2022automated} and teaching assistance \cite{zylich2020exploring}. Still, to the best of our knowledge, our study presents the first evaluation of a QA system designed specifically for answering university students' questions about their study programs and requirements.

\section{Dataset Creation}
\label{chapter:corpus}
The dataset consisted of study program descriptions and requirements by scraping and cleaning 72 university study program descriptions to be used a representative of a domain-specific dataset. We also generated an evaluation set consisting of question-answer and Context triplets. 

The origin of documents used for corpus creation are the official websites of a large technical university in Germany. Every document describes a certain study program at the bachelor's or the master's level, covering various fields like natural sciences, engineering, and management. The documents contain information such as the name and overview of the study program, admission prerequisites, coursework, examinations, graduation requirements, credits earned, study fees, deadlines, etc. All the study program descriptions were scraped from the official university website with the Python library \textit{BeautifulSoup} and saved in a JSON format. In total, there were 72 different study programs, with parallel text written in both German and English.

Once the websites were scraped and a corpus of documents for each study program was created, a representative test set of questions that covered university students' information needs was constructed. This process was done semi-automatically by combination of GPT-4 and human-in-the-loop by scrapping frequently asked questions and also generation some new questions with the help of GPT-4. The whole process resulted in 500 QA pairs that served as the evaluation set. To ensure the quality and relevance of the generated QA pairs, the two authors manually went over all the questions and filtered the dataset down to 200 QA pairs. The questions that were kept aimed to cover a diverse range of students' personalized information needs \cite{https://doi.org/10.1002/asi.24234}. Additionally, we consulted with students from both bachelor's and master's levels and inquired about the most common questions they have asked over the years related to their program's regulations.  All 200 answers were checked and any observed errors that stem from GPT generation were manually corrected.

The questions cover the most common information needs of the students, focusing on factual knowledge such as duration of the study program, total credits required to graduate, and admission requirements to programs; but also the more complex questions that require reasoning over multiple parts like future job prospects, knowledge gained in the program, or the difference in scope between to programs. Some example questions are displayed in \autoref{tab:sample-questions}.

\begin{table}
 \caption{Selected questions from the curated study program dataset.}
\begin{tabular}{p{14cm}l}
\midrule
1) What are the tuition fees for international students in the Aerospace Master of Science (M.Sc.)?\\
2) How do I apply for the Master’s program in Management if I have an undergraduate degree from outside the EU?\\
3) How many ECTS points do I need for Mathematics in Data Science?\\
\hline
    \end{tabular}
    \label{tab:sample-questions}
\end{table}

\section{RAG Framework}
\label{chapter:methodology}

This Framework is designed to explicate both the theoretical underpinnings and practical implementations of the RAG framework, encompassing both open-sourced and closed-sourced models in English and German. Key modules such as the Multi-Query module (Generation Phase), Child-Parent Retriever and Ensemble Retriever (Retrieval Phase), and In-context-learning (Generation Phase) are discussed to highlight their roles in enhancing the framework's performance. Each module's inclusion reflects strategic choices made to enhance the framework's efficiency and effectiveness in generating contextually relevant answers.

\subsection{Models}

We incorporated both open-source models such as \textbf{Llama 2} \cite{touvron2023llama} including \texttt{llama2:7b-chat} and \texttt{llama2:13b-chat}, and \textbf{Mistral AI} \cite{jiang2023mistral} specifically the \texttt{mistral:7b-instruct} version. Given the success of closed-source \textbf{OpenAI models}, we included \texttt{gpt-3.5-turbo-1106} and \texttt{GPT-4-0125-preview} in our RAG framework due to their performance in language comprehension, creativity, and complex query handling.

\subsection{Optimization Modules}

\subsubsection{Pre-retrieval Phase}
We follow a two-step approach for this such that we first try to find the respective study program and then the topic (Costs, Required Language Proficiency, Type of Study, etc) matching the student's query. We Identify both through an LLM-based filtering mechanism, that is later used in tandem with the Retriever.  The selection of a study program is executed by correlating the user's query with a predefined list of programs. Combining the user query and the possible study programs in the following prompt template yields the answer. 
    \begin{lstlisting}[frame=tb, postbreak={} , breakindent=0pt, breaklines=true,  autogobble=true, basicstyle=\footnotesize\ttfamily]
    System: The following will be a query by a student about study programs of a Technical University. I want you to output the study program the student is having questions about. Only output the study program!
    Context: Please note, that we only need the study program. Do not self-reference, comment or give any notes. Only output the study program. Here are the possible study programs: {listed_root_level_keys}
    Human: How many ECTS points do I need for Mathematics in Data Science?
    Response: 
    \end{lstlisting}
We match the generated study program name with the exact study program names embedded in our \textit{vectorstore} using cosine similarity. 
Upon obtaining the study program from the initial query, the next step is to put the user's query in a structured template designed to direct the model's focus toward generating topics relevant to the study program. 
We again match the predicted topic with one of the embedded chunks using cosine similarity.

\subsubsection{Retrieval Phase}

We use \textbf{Multi-Query} \footnote{\url{https://docs.llamaindex.ai/en/latest/examples/retrievers/reciprocal_rerank_fusion/}}, a module utilized to generate variations of the user's original question, aiming to explore different formulations that could prompt the LLM to provide a broader range of topics as context during the Generation Phase. In our implementation, the study program remains constant; only the user's question is rephrased. We use a prompt-based approach with GPT-4 to generate similar queries. This strategy ensures we can elicit more detailed or varied information relevant to the same study program without altering its core focus. 
Secondly, through the \textbf{Child-Parent Retriever} \footnote{\url{https://python.langchain.com/docs/modules/data_connection/retrievers/parent_document_retriever/}}, we distinguish between two types of vector stores: the "memory store" for child chunks and the vectorstore for parent chunks, each tailored for different granularities of data. The memory store ingests data segmented into 300-character chunks, while the vectorstore processes larger 1500-character chunks. All embeddings for these vector stores are generated using the \texttt{all-MiniLM-L6-v2} model, ensuring consistency across the embedding process. The challenge is to reconcile the need for small document chunks, which allow for more accurate embeddings, with the requirement for maintaining sufficient document size to ensure context is preserved. Each child chunk is assigned a unique ID, and corresponding parent chunks maintain a list of these IDs. During retrieval, the process identifies a relevant child chunk in the memory store, retrieves its unique ID, and then locates the parent chunk containing that ID in its list. This approach ensures that while the embedding of small chunks accurately reflects their content, the broader context is not lost, as parent chunks provide a comprehensive view necessary for subsequent processing steps. Lastly, the \textbf{Ensemble Retriever} integrates two retrieval methods: BM25, which prioritizes documents based on direct term matches, and cosine similarity, which focuses on the contextual relevance between texts. The Ensemble Retriever combines the BM25\cite{Robertson2009BM25} retrieval function with cosine similarity, offering adjustable weighting between the two methods to optimize search outcomes However, during our experiments we assign equal weight to both.



\subsection{Generation Phase}

We use In-context learning, which allows LLMs to learn new tasks on run-time through examples passed as part of the input prompt. We explore three-shot learning in our experiments to evaluate if it would improve the performance of an RAG framework. A considerable challenge was selecting three Question-Answer pairs that exhibit a broad semantic diversity to prevent model bias towards the domains of the examples provided.

\subsection{Evaluation}
The growing importance of RAG-based applications requires a detailed evaluation method that separates Retrieval Quality and Generation Quality. 
\subsubsection{HIT Rate}
Hit Rate measures the proportion of queries for which the correct answer is found within the top-5 retrieved documents. Scoring a hit implies not only detecting the correct document (study program) but also the topic of interest within that document. 

\subsubsection{Reference-based Evaluation}
We also employed some reference-based evaluation metrics like ROUGE\cite{lin2004rouge} metric, which compares generated answers with a reference text, and assesses content overlap through n-grams and other sequences. Unfortunately, ROUGE's approach may miss the mark in recognizing semantically accurate or contextually fitting answers that don't exactly replicate reference phrasing, pointing to a need for metrics that evaluate semantic and contextual accuracy in RAG systems. Hence, we also calculate the contextual similarity between the reference and generated answer through BERTScore\cite{zhang2020bertscore}
\subsubsection{LLM-based Evaluation}
State-of-the-art research revolves around evaluating LLM-generated text across multiple features \cite{zhang2023benchmarking,yang2023empower}. We explore the concept of using LLM as an evaluator as described in the G-eval\cite{liu2023geval} on a scale of 1 - 5 across Relevance, Coherence, Fluency, and Faithfulness.

\subsubsection{RAG Confusion Matrix}

The traditional Confusion Matrix is a fundamental tool in machine learning for evaluating the performance of classification models. In the context of RAG systems, the Confusion Matrix concept is adapted to assess both the retrieval and generation phases of these models. We derive the confusion matrix for the same features as the ones evaluated by LLM. We consider an answer to be acceptable if the evaluation score assigned is above a threshold. Thus, the RAG Confusion Matrix evaluates: 
\newline\textbf{True Positive (TP):} Relevant do is retrieved and an acceptable response is generated, \newline\textbf{True Negative (TN):} Relevant document is not retrieved and an acceptable response is not generated
\newline\textbf{False Positive (FP):} Relevant document is not retrieved and an acceptable response is generated
\newline\textbf{False Negative (FN):} Relevant document is retrieved and an acceptable response is not generated

\section{Results and Discussion}
\label{chapter:evaluation_results}
Given the information related to a study program, we asked LLM to generate question-answer pairs from it. After manual curation, 200 pairs were used for evaluation involving 96 different experiments, using a variety of LLMs, in both German and English.\footnote{Unless specified otherwise, we used all methods in their default configurations.} 

\subsection{HIT Rate}

We measure the effectiveness of a model in extracting relevant textual information from a dataset by calculating the Hit Rate for each experiment on a subset of 82 samples. \autoref{Table:Hit_Rate} shows that any module iteration excluding the Multi-Query feature performs significantly worse than those that incorporate it, highlighting the critical role of Multi-Query in enhancing retrieval effectiveness. Furthermore, only the closed-sourced models show competence in the German language while all open-source models except Mistral struggle with it.

\begin{table}[h]
\small
\centering
\caption{Hit Rate over all RAG configurations where er = Ensemble Retriever, cpr = Child-Parent-Retriever, icl = In-Context-Learning, mq = Multi-Query.}
\label{Table:Hit_Rate}
\begin{tabular}{c c c c c c c c c c }

{Model} & & {er} & {cpr} & {icl} & {icl-er} & {mq-er} & {mq-cpr} & {mq-cpr-icl} & mq-icl-er\\ 
\midrule \multirow{2}{*}{ Llama 2~7B } & de & 8.64 & 8.64 & 8.64 & 7.41 & {13.58} & {13.58} & {13.58} & {13.58} \\
 & en & 43.21 & 43.21 & 43.21 & 43.21 & {53.09} & {53.09} & {53.09} & {53.09} \\
 \hline \multirow{2}{*}{ Llama 2~13B } & de & 28.40 & 28.40 & 28.40 & 28.40 & {34.57} & {34.57} & {34.57} & {34.57} \\
 &  en & 50.62 & 50.62 & 50.62 & 50.62 & {55.56} & {55.56} & {55.56} &{55.56} \\
\hline \multirow{2}{*}{ GPT 3.5}  & de & 56.79 & 56.79 & 58.02 & 55.56 & 66.67 & 64.20 & {70.37} & 65.43 \\
 &  en & 44.44 & 41.98 & 41.98 & 43.21 & {46.91} & 44.44 & 45.68 & 45.68 \\
\hline \multirow{2}{*}{ GPT-4} & de & 61.73 & 61.73 & 61.73 & 61.73 & \textbf{69.14} & {66.67} & 65.43 & {66.67} \\
 &  en & 65.43 & 66.67 & 66.67 & 66.67 &  \textbf{75.31} & 72.84 & 72.84 & 72.84 \\
\hline \multirow{2}{*}{ Mistral~7B } & de & 39.51 & 39.51 & 39.51 & 39.51 & 48.15 &  \textbf{51.85} &  \textbf{51.85} & 49.38 \\
 &  en & 51.85 & 51.85 & 53.09 & 51.85 & 56.79 & 56.79 & 56.79 & 56.79 \\
\hline
\end{tabular}
\end{table}

\begin{table}[]
\small
\centering
\caption{Metric: Faithfulness evaluated by GPT-4 on a scale of 1 - 5.}
\label{Table:AVG_Table_Faithfulness}
\begin{tabular}{c c c c c c c c c c c }
{Model} & & & {er} & {cpr} & {icl} & {icl-er} & {mq-er} & {mq-cpr} & {mq-cpr-icl} & mq-icl-er\\ 

\midrule \multirow{4}{*}{ Llama 2~7B} & \multirow{2}{*}{de} &  {Match} & {2.7} & {2.6} & {2.4} & {2.6} & {4.0} & {3.6} & {2.9} & {3.6} \\
&                                                             & {No Match}& {1.5} & {1.6} & {2.1} & {2.0} & {1.9} & {2.0} & {3.1} & {3.1}\\
 & \multirow{2}{*}{en}                                      & {Match} & \textbf{4.7} & {3.2} & {3.2} & {2.1} & {4.5} & {3.2} & {3.6} & {3.1} \\
&                                                          &{No Match}& {3.5} & {2.5} & {2.3} & {2.8} & {2.6} & {2.4} & {3.2} & {2.5} \\
 
 \hline \multirow{4}{*}{ Llama 2~13B } & \multirow{2}{*}{de} & {Match} & {4.4} & {3.1} & {3.5} & {4.2} & {3.0} & {2.8} & {2.7} & {3.3} \\
 &&                                                         {No Match}& {3.3} & {3.1} & {2.4} & {3.6} & {2.0} & {3.3} & {2.8} & {2.9} \\
 &  \multirow{2}{*}{en}                                      & {Match}& {3.9} & {3.2} & {2.4} & {3.4} & {3.8} & {2.7} & {3.2} & {4.6}\\
&                                                         &{No Match}& {2.6} & {2.6} & {1.4} & {3.0} & {2.7} & {2.1} & {2.3} & {2.7} \\

\hline \multirow{4}{*}{ GPT 3.5}  & \multirow{2}{*}{de} &{Match} & {4.2} & {3.2} & {2.9} & {3.8} & {3.9} & {3.2} & {3.6} & {3.9} \\
&                                                       &{No Match}& {2.9} & {3.2} & {4.0} & {3.1} & {2.7} & {2.6} & {2.9} & {2.1}  \\
 &  \multirow{2}{*}{en}                                 & {Match}& {4.3} & {2.8} & {2.5} & {3.1} & {3.6} & {3.4} & {3.1} & {4.0}\\
&                                                        &{No Match}& {2.1} & {1.9} & {3.5} & {2.7} & {1.8} & {1.5} & {2.5} & {1.9}  \\

\hline \multirow{4}{*}{ GPT-4} & \multirow{2}{*}{de} & {Match}& {4.4} & {2.6} & {3.3} & {3.9} & \textbf{4.8} & {2.5} & {3.3} & {4.0}\\
 &                                                       &{No Match}& {3.2}& {4.0}& {3.8}&{3.0}& {3.4}& {3.1}& {3.1} & {3.2} \\
 &  \multirow{2}{*}{en}                                    & {Match}& {3.4} & {3.0} & {3.3} & {4.3} & \textbf{5.0} & {3.6} & {3.5} & {3.8} \\
&                                                        &{No Match}& {2.3} & {2.6} & {2.7} & {2.7} & {3.3} & {2.2} & {2.7} & {2.9} \\

\hline \multirow{4}{*}{ Mistral~7B }  & \multirow{2}{*}{de} & {Match} & \textbf{4.7} & {2.9} & {3.2} & {4.5} & {4.4} & {4.0} & {2.6} & {4.1}\\
&                                                          &{No Match}& {2.9} & {2.7} & {3.4} & {3.9} & {3.3} & {3.8} & {4.5} & {4.3}  \\
 & \multirow{2}{*}{en}                                     &  {Match} & {3.7} & {2.4} & {3.5} & {4.4} & {4.6} & {2.3} & {3.2} & {4.2}\\
&                                                         &{No Match}& {2.5} & {2.5} & {2.2} & {3.3} & {3.5} & {2.7} & {2.2} & {2.3}  \\
\hline
\end{tabular}
\end{table}

\subsection{LLM as an evaluator}

LLM-based evaluation metrics are known to perform on par with human evaluation \cite{afzal2024HIL}. Since it is costly to perform human evaluation over such a wide range of heuristics, we employ GPT-4 for a proxy evaluator to narrow down the search space and hence find the optimal configurations for RAG. To keep our analysis practical and cost-effective, we included a sample of 20 answers, half with matches to the correct context (Match) and the other half to incorrect context (No Match).

\subsubsection{Faithfulness over others}
For our use case, Faithfulness holds slightly more weight than other metrics. As seen in \autoref{Table:AVG_Table_Faithfulness}, configurations incorporating the Ensemble Retriever demonstrated superior performance, with several configurations achieving average scores above 4, in both English and German. The most consistently high-performing combinations involve the Multi-Query and Ensemble Retriever, with or without the addition of ICL.

\subsubsection{Confusion Matrix}

\begin{table}[!h]
\small
\centering
\caption{RAG confusion matrix across Faithfulness, Relevance, Coherence, and Fluency of top 3 configuration selected while using Faithfulness as a heuristic with a Threshold of 5. Match/No Match means that Correct/Incorrect context was retrieved.}
\vspace{0.2cm}
\label{Table:Top_TP_fth_Rel_Coh_Flu_Scores}
\vspace{-12pt}

\begin{tabular}{c|c c|c c|c c}

\multicolumn{1}{c}{} & \multicolumn{2}{c}{\textbf{ChatGPT4 (en-mq-er)}} & \multicolumn{2}{c}{\textbf{Llama2 7B (en-er)}} & \multicolumn{2}{c}{\textbf{Mistral 7B (en-mq-er)}} \\
Match & Correct & Incorrect & Correct & Incorrect & Correct & Incorrect \\
\midrule
\multicolumn{7}{c}{Faithfulness} \\
\midrule

1 & $10(\mathrm{TP})$ & $0(\mathrm{FN})$ & $9(\mathrm{TP})$ & $1(\mathrm{FN})$ & $9(\mathrm{TP})$ & $1(\mathrm{FN})$ \\
0 & $5(\mathrm{FP})$ & $5(\mathrm{TN})$ & $4(\mathrm{FP})$ & $6(\mathrm{TN})$ & $5(\mathrm{FP})$ & $5(\mathrm{TN})$ \\

\midrule
\multicolumn{7}{c}{Relevance} \\
\midrule

1 & $7(\mathrm{TP})$ & $3(\mathrm{FN})$ & $7(\mathrm{TP})$ & $3(\mathrm{FN})$ & $9(\mathrm{TP})$ & $1(\mathrm{FN})$ \\
0 & $4(\mathrm{FP})$ & $6(\mathrm{TN})$ & $4(\mathrm{FP})$ & $6(\mathrm{TN})$ & $5(\mathrm{FP})$ & $5(\mathrm{TN})$ \\

\midrule
\multicolumn{7}{c}{Coherence} \\
\midrule

1 & $10(\mathrm{TP})$ & $0(\mathrm{FN})$ & $8(\mathrm{TP})$ & $2(\mathrm{FN})$ & $9(\mathrm{TP})$ & $1(\mathrm{FN})$ \\
0 & $3(\mathrm{FP})$ & $7(\mathrm{TN})$ & $5(\mathrm{FP})$ & $5(\mathrm{TN})$ & $3(\mathrm{FP})$ & $7(\mathrm{TN})$ \\

\midrule
\multicolumn{7}{c}{Fluency} \\
\midrule
1 & $10(\mathrm{TP})$ & $0(\mathrm{FN})$ & $8(\mathrm{TP})$ & $2(\mathrm{FN})$ & $10(\mathrm{TP})$ & $0(\mathrm{FN})$ \\
0 & $7(\mathrm{FP})$ & $3(\mathrm{TN})$ & $9(\mathrm{FP})$ & $1(\mathrm{TN})$ & $6(\mathrm{FP})$ & $4(\mathrm{TN})$ \\
\end{tabular}
\end{table}
The proposed confusion matrix helps us decouple LLM's generation abilities from the retriever and provides a deeper insight into both modules of RAG. Such an analysis helps us understand which parts of the retriever or generator needs to be improved. Based on \autoref{Table:AVG_Table_Faithfulness}, we compute the RAG confusion matrix on the best-performing configuration as per Faithfulness where "Match" means the correct context was retrieved by the retriever and "no Match" means that an incorrect context was retrieved. For each evaluation feature from Relevance, Coherence, Fluency, and Faithfulness we set a threshold of 5 for our confusion matrix. We consider it a correct response if the LLM (GPT-4) rates the answers a score of 5. The Confusion matrix summarized in \autoref{Table:Top_TP_fth_Rel_Coh_Flu_Scores} provides interesting insights regarding high False Positive (FP) values. As expected, the Faithfulness scores for No Match samples are quite low since the incorrect context was provided to the LLM for generation. This suggests that LLMs tend to be much more Faithful if they are provided with the correct context; some more than the other though.

\subsection{Further Enhancements and Evaluation of RAG}
\label{improved RAG}
 Having found the top 3 optimal RAG configurations in the previous set of experiments, we tried to optimize them. For the retriever part, our optimization efforts revolved around the filtering steps in the pre-retrieval phase. We discovered that allowing the LLM to freely predict the study program instead of constraining it to a predefined list helped increase the hit rate. We compute LLM-based evaluation scores, ROUGE, and BERTScores in \autoref{Table:Optimized_AVG_Table} which shows the leading performance of GPT-4. ROUGE and BERTScore reflect GPT-4's its strong ability to generate responses that closely align with the reference texts, both in detail and semantic relevance, indicating proficiency in producing contextually accurate and engaging content. In general, all models produce somewhat fluent and coherent answers with some discrepancies in Faithfulness and Relevance. An answer can be faithful without being relevant but not the other way around. An example of this can seen in Mistral 7B which has a higher Faithfulness score but comparatively low Relevance scores. Due to the limited data sample, definitive conclusions are challenging to draw. GPT-4, however, maintained similar performance levels across all metrics. This consistency is noteworthy, especially when considering the divergence in evaluations between human evaluators and GPT-4 for the optimized RAG frameworks. This final interpretation aligns well with the findings that the performance of Llama 2 and Mistral dropped compared to the original RAG framework.

\subsubsection{Correlation between LLM-based and Human Evaluation}

We provide comparison of the evaluation scores between the normal/improved framework with GPT-4/human evaluation in \autoref{Table:Optimized_AVG_Table} calculated on a subset of 30 samples. Llama 2 and Mistral didn't align with human preferences as well as GPT-4. This could indicate that GPT-4, having processed vast amounts of data, can present information in a manner more appealing to humans.

\begin{table}[h]
\small
\centering
\caption{Faithfulness, Relevance, Coherence, and Fluency scores with mq-icl-er where "x" represents evaluation by GPT-4 and "x/x" represent "evaluation by GPT-4 / Human" on English Dataset. ROUGE and BERTScore are computed using the reference answers. Retriever+ refers to the optimized version explained in Section\ref{improved RAG}.}
\vspace{-7pt}
\label{Table:Optimized_AVG_Table}
\begin{tabular}{c c  c c c c c}
&  Retriever & Retriever+ & {Retriever} & Retriever+ & Retriever & Retriever+ \\
&  Llama 2~13B & Llama 2~13B & {GPT-4} & {GPT-4} & Mistral~7B& Mistral~7B \\
\hline
\multicolumn{7}{c}{ {Faithfulness} }\\
\hline  Match & 
 3.2 & 3.2/2.7 
 & 3.4 & \textbf{4.4/4.3} 
 & \textbf{4.5} & 4.2/3.4 \\ 
No Match & 
 2.0 & 2.1/2.2 
 & 2.9 & 3.5/3.3 
 & 2.5 & 3.3/2.6 \\ 
 \hline
\multicolumn{7}{c}{ {Relevance} }\\
\hline Match & 
 3.5 & 3.3/3.0 
 & 3.4 & \textbf{4.4/4.1} 
 & \textbf{4.6} & 3.9/2.9 \\ 
No Match & 
 2.7 & 2.7/2.3 
 & 2.6 & 3.3/3.7 
 & 2.5 & 3.4/2.1 \\ 
\hline

\multicolumn{7}{c}{ {Coherence} }\\
\hline Match & 
3.8 & \textbf{4.5}/3.9 
& 3.4 & 4.4/3.8 
& \textbf{4.5} & \textbf{4.6}/\textbf{4.0} \\ 
 No Match & 
2.5 & 2.6/3.4 
& 2.4 & 3.2/\textbf{4.4} 
& 2.9 & 3.2/3.0 \\ 
\hline
\multicolumn{7}{c}{ {Fluency} }\\
\hline  Match & 
 4.2 & 4.5/4.8 
& 4.5& \textbf{4.9}/\textbf{4.8} 
& \textbf{4.9} & \textbf{4.9}/\textbf{4.9} \\ 
No Match & 
4.4 & 4.4/4.6 
& 4.1 & \textbf{4.8}/\textbf{5.0} 
& 4.4 & 4.7/4.7 \\ 

\hline
\multicolumn{7}{c}{ {ROUGE} }\\
\hline  
Match & {0.239} & {0.306} 
& \textbf{0.538}& {0.484} 
& {0.314} & {0.288} \\ 
No Match & {0.184} & {0.278} 
& {0.336} & \textbf{0.434} 
& {0.256} & {0.196} \\ 

\hline
\multicolumn{7}{c}{ {BERTScore} }\\
\hline  
Match & {0.881} & {0.883} 
& \textbf{0.930}& {0.910} 
& {0.887} & {0.893} \\ 
No Match & {0.858} & {0.873} 
& {0.880} & \textbf{0.904} 
& {0.865} & {0.863} \\ 
\end{tabular}
\end{table}

\section{Conclusion and Future Work}

During our experiments, we investigated various optimizations over a RAG framework while employing Faithfulness as the most prominent evaluation criterion. We evaluated the effectiveness of the Multi-Query, Child-Parent-Retriever, Ensemble Retriever, and In-context learning. Additionally, we introduce two novel ideas; first, the usage of LLM-based filtering in the pre-retrieval phase that aids in the identification of the correct study program \& topic, and second the adaptation of the confusion matrix for RAG evaluation. Our initial experiments show a significant improvement in Hit Rate with the incorporation of Multi-Query techniques. Conversely, as per the LLM-based evaluation of the generated answer, Ensemble Retriever showed remarkable effectiveness in terms of Faithfulness, particularly when paired with the Multi-Query. While GPT-4 is a clear winner, there are notable competitors such as open-source Llama 2 13B for the Retrieval Phase, leveraging Multi-Query, In-Context-Learning, and Ensemble Retriever. For the Generation Phase, the Original Mistral 7B with the same module iteration outperforms other models. Lastly, we found discrepancies between Human and LLM-based evaluation as the human annotators as the humans almost always rated the answer lower than the LLM rating. Future work could include the incorporation of an even wider array of open-source models with larger sizes, to facilitate a fairer comparison to a model as large as GPT-4. 

\bibliographystyle{ACM-Reference-Format}
\bibliography{mybibfile}
\vspace{-35pt}

\end{document}